# Residual and Attentional Architectures for Vector-Symbols

Wilkie Olin-Ammentorp
Department of Medicine,
University of California, San Diego
San Diego, CA, USA
wolinammentorp@ucsd.edu

Maxim Bazhenov
Department of Medicine, Institute for Neural Computation,
University of California, San Diego
San Diego, CA, USA
mbazhenov@ucsd.edu

## ABSTRACT

Vector-symbolic architectures (VSAs) provide methods for computing which are highly flexible and carry unique advantages. Concepts in VSAs are represented by 'symbols,' long vectors of values which utilize properties of high-dimensional spaces to represent and manipulate information. In this new work, we combine efficiency of the operations provided within the framework of the Fourier Holographic Reduced Representation (FHRR) VSA with the power of deep networks to construct novel VSA based residual and attention-based neural network architectures. Using an attentional FHRR architecture, we demonstrate that the same network architecture can address problems from different domains (image classification and molecular toxicity prediction) by encoding different information into the network's inputs, similar to the Perceiver model. This demonstrates a novel application of VSAs and a potential path to implementing state-of-the-art neural models on neuromorphic hardware.

## 1 Introduction

Vector-symbolic architectures (VSAs) have been undergoing renewed interest due to their potential use as a 'common language' for neuromorphic computing [13]. Each VSA uses a long, 'hyperdimensional' symbol to represent information. Sets of these symbols can then be manipulated and reduced via algebraic operations termed 'binding' and 'bundling.' These operations produce composite symbols encoding complex data structures such as sets, images, graphs, and more [11, 16]. A VSA also includes a 'similarity' operation which provides a metric to capture how closely two symbols relate to one another.

Researchers have applied VSA-based techniques to a variety of problems, such as similarity estimation, classification, analogical reasoning, and more [12]. Increasingly, the approaches involving integration of neural networks to transform VSA symbols has been getting attention in the field. The neural networks can play a role of mapping symbols between different domains, for instance, converting an 8-bit color image into a symbol. Neural networks may also convert between symbols in different informational domains, such as converting a symbol representing an image into one representing a label [17]. However, as is well known from the larger field of deep neural network research, changing the architecture of a neural network is often required to allow a network to satisfactorily solve a given problem [5]. More 'difficult' problems can require networks which can be scaled up efficiently and effectively. In computer vision, this has often been done via the use of convolutional layers [15]. However, we suggest that convolutional layers are not well-suited to processing VSA symbols on two grounds: firstly, that VSAs are inherently designed to provide distributed representations of information [20]. This conflicts with the design of convolutional networks to extract local correlations from inputs via the use of kernels [5]. Secondly, convolutional networks often change scale by using differently-shaped kernels and feature maps at each layer. This contrasts with VSAs, which maintain the dimensionality of a symbol at each processing step. This feature is potentially an advantage of VSAs which make them good candidates for neuromorphic hardware, which would not have to be designed to account for primitives which can change shape during processing. For these reasons, we do not utilize convolutional methods for processing VSA symbols and instead focus on applying fully-connected layers and attention-based methods.

In this work, we demonstrate how two advances – residual and attentional architectures – may be naturally integrated into a neural network processing VSA symbols. Specifically, this integration is done within the domain of the Fourier Holographic Reduced Representation (FHRR) VSA [19], allowing these networks to remain compatible with potential neuromorphic hardware. We demonstrate that by incorporating these advances to create attention-based modules for processing symbols, these architectures provide a powerful and scalable method for learning complex mappings.

### 1.1 Fourier Holographic Reduced Representation

In this work, we adapt the use of the FHRR VSA for all symbolic representations, as it can be efficiently implemented with deep learning frameworks, performs well empirically, and has unique links to spiking neural networks [4, 21]. In the FHRR VSA, each element of a symbol represents an angular value. We normalize these values by $\pi$ to represent angles on the domain [-1,1]. Angular values can therefore be converted into a complex number via Euler's formula $e^{i\pi x} = \cos \pi x + i \sin \pi x$, where $i$ represents the complex unit and $x$ is a vector of radian-normalized angles.

To measure the similarity (abbreviated to 'sim.' in formulas) between two FHRR vector-symbols $a$ and $b$ with $n$ elements, the mean value of the cosine of element-wise angular differences is found (Eqn. 1). This means two symbols can be identical (similarity values of 1.0) or unrelated (similarity values close to 0.0). Symbols in the FHRR can also be 'opposing' (similarity values of -1.0), but this feature is not used in this work.

$$sim.\,(a, b) = \frac{1}{n} \sum_{i=1}^{n} \cos{(a_i - b_i)}$$

(Eqn. 1)

To produce one symbol which is maximally similar to a set of inputs, the *bundling* operation (+) is used. A set of $m$ input vectors with dimensionality $n$ may be stacked into a single $m \times n$ matrix $A$. This matrix of radian-normalized angles is converted to complex values and summed along its first axis. The angle of the resulting row vector is taken to produce a single new symbol from the input set (Eqn. 2).

$$+(A) = angle\left(\sum_{j=1}^{m} \exp{(i\pi A_{j,:})}\right)$$

(Eqn. 2)

*Binding* (×) can be used to combine concepts represented by different symbols into a new symbol dissimilar to its inputs. In the FHRR, this is done simply by 'rotating' the angles in an input symbol $a$ by those in a separate 'displacement' vector, $b$ . *Fractional* binding can be accomplished by including a power $p$, which multiplies the amount by which the displacement vector rotates the input. This can be used to encode continual values within a vector-symbolic representation (Eqn. 3). Normal binding uses a power of 1.0.

$$\times (a, b, p) = (a_i + p \cdot b_i + 1) \% 2 - 1$$

(Eqn. 3)

We demonstrate that using these operations and their sub-components (e.g., matrix multiplication in Eqn. 2), trainable and effective neural models with deep and attentional mechanisms can be created.

### 1.3 Generalized Bundling

The bundling function may be rewritten in a more general form which can allow it to serve as the basis for a trainable neural layer. As previously, $m$ input vector-symbols with $n$ angles each are taken as an input, represented as an $m \times n$ matrix $A$ which is converted into the complex domain. However, by including an $n \times o$ matrix of projection weights $W_p$, the complex values can be projected into a new $m \times p$ output space. To reduce or expand these projected values, an $r \times m$ set of reduction weights $W_r$ can also be included (Eqn. 4). If $W_r$ is an $1 \times m$ matrix of ones and $W_p$ is an $n \times n$ identity matrix, this "generalized" bundling function reduces back to the previous case (Eqn. 2). If instead $W_r$ is an $m \times m$ identity matrix and $W_p$ is a trainable $n \times p$ matrix, generalized bundling can be used as a neural layer with a non-linear activation function which produces an $m \times p$ output from an $m \times n$ input $A$. We have previously demonstrated the use of this layer to produce effective neural networks which can be executed via spiking dynamics [18].

$$+(A, W_r, W_p) = angle(W_r \cdot \exp(i\pi A) \cdot W_p)$$

(Eqn. 4)

This generalized bundling function is used as the basis of all fully-connected layers in this work. Previously, we have also termed this the 'phasor' activation function, but here refer to it as the generalized bundling function to emphasize its close relationship to the vector-symbolic operation.

## 2 Results

### 2.1 Residual Layers

When layers using generalized bundling as a neural layer are stacked into deeper networks, the same issue is found as when traditional activation functions (e.g. ReLU) are used: the conditioning of deeper networks becomes poor, and they become difficult or impossible to train in practice. This was addressed in traditional networks by the introduction of residual blocks, which modify one or more layers so that when using initial weights, their output approximates the identity function [8]. This is achieved simply by using a 'skip' connection to add the input values of a block to its outputs, as the output of a layer with random weights and most common activation functions is approximately zero.

Adapting generalized bundling (PB) layers to form residual blocks encounters a practical difficulty which we address here. Firstly, a zero-centered random initializer (Gaussian, uniform, or otherwise) in a PB layer will weight and sum a set of inputs to produce a complex value which is centered on the origin of the complex plane (Figure 1a). As a result, the generalized bundling activation function operating with initial random projection weights will produce values which are not normally distributed around zero, but uniformly distributed random angles. This violates one of the requirements for a residual block: that a layer will initially produce values narrowly centered on zero.

To address this, a complex bias can be added to the generalized bundling function. This bias can shift the distribution of initial complex values. In this work, the bias is set to move the origin of projected values from $0 + 0i$ to $1 + 0i$ (Figure 1a). Given this change, the distribution of angular values produced by generalized bundling becomes normally centered around zero (Figure 1b). With this change, a skip connection operating on the output of a PB layer can approximate the identity function. Additionally, we make the substitution of the VSA 'binding' operation on the output of the PB layer to replace addition, as the former restricts the output to be on the FHRR's domain and allows the computation to remain compatible with hardware which fully implements VSA operations but not arbitrary mathematical ones. These changes produce a residual block which solely employs operations used within the VSA.

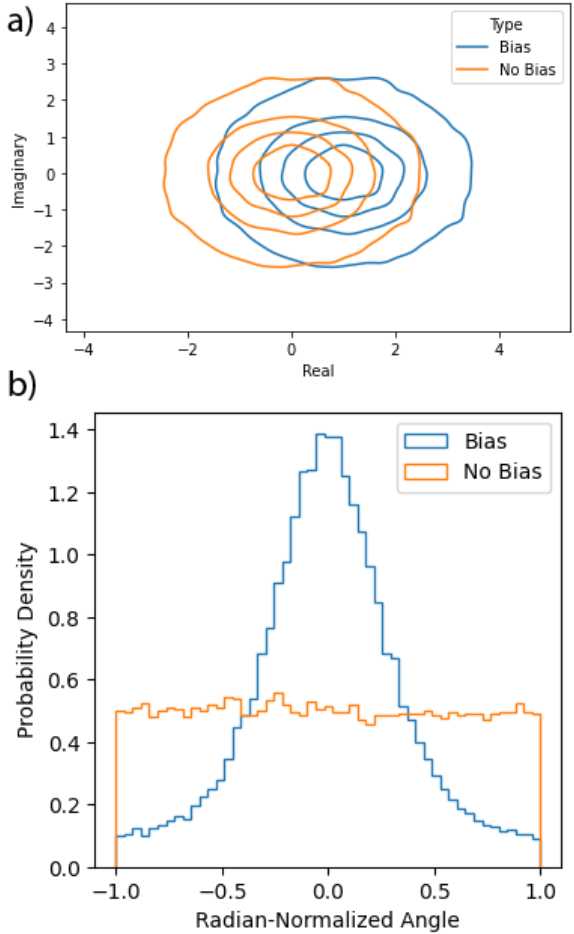

**a)**

**b)**

**Figure 1: Using a standard weight initialization scheme leads to a projection of values centered around the origin of the complex plane. By offsetting this projection with a bias (a), the angles of these complex values are shifted from a uniform distribution to a normal one centered on zero (b).**

We validate this approach by using a simple image classification task. Images from the FashionMNIST dataset are transformed into FHRR symbols by Gaussian random projection and Layer Norm [1, 23]. These symbols are processed via successive autoencoding multi-layer perceptron (MLP) blocks. Each block contains one hidden layer twice the width of a symbol ($2n$) and an output layer which reduces it back to the symbol dimensionality $n$, with both layers' outputs calculated via the generalized bundling function. These blocks may be followed by a skip connection to create a residual network. The final output symbol is compared to a set of random symbols representing each image class, and the class the output is most similar to is taken as the predicted label (Figure 2). Loss is calculated by comparing the similarity of the network's output to the symbol corresponding to the correct class.

The improved conditioning induced by applying a binding-based skip connection allows deep networks to become trainable (Figure 3). A network using PB layers and VSA residual blocks with 24 total layers reaches 85.8% test accuracy on the FashionMNIST test split. However, if the skip connections are removed the model does not exceed chance levels of classification accuracy.

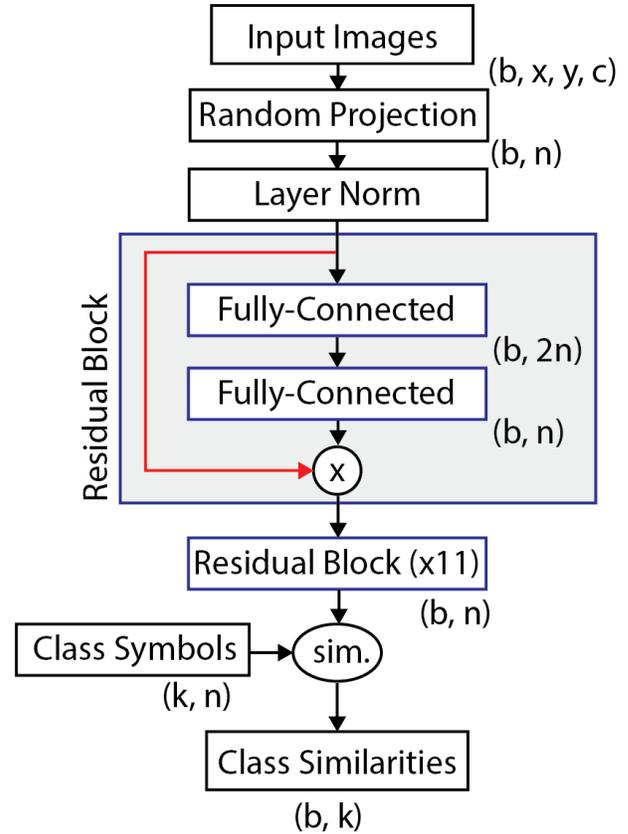

**Figure 2: Illustration of the deep architecture used to test network trainability. Layers in the architecture which produce a new output shape are labeled, with this new shape in parentheses below. First, input images are flattened and projected to the dimensionality of the vector space being used (n) and normalized into [-1, 1] using a LayerNorm. These symbols then pass through residual blocks consisting of two fully-connected layers based on the generalized bundling function and a skip connection (red). The vector-symbolic similarity of the blocks' outputs is compared to the class symbols, yielding a prediction for each image's class.**

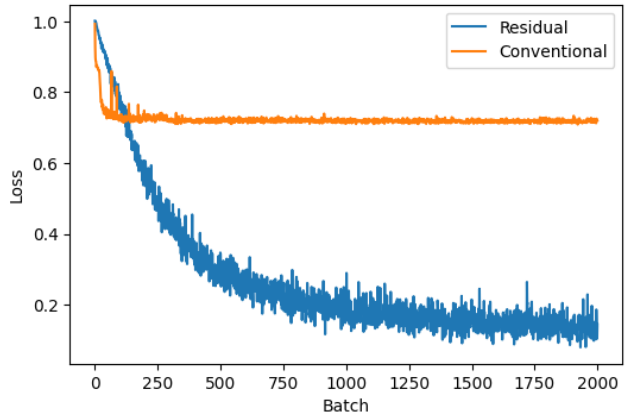

**Figure 3: Adapting residual blocks into an FHRR-based network with 24 layers improves the conditioning of the network, allowing it to become trainable (blue) in comparison to its counterpart without binding-based skip connections (orange).**

## 2.2 Attention Mechanisms

The cornerstone of many recent advances in natural language processing (NLP) tasks is "attention," the ability to compute intermediate representations of inputs and calculate how these representations relate to one another and should be used to adjust the information which is passed downstream [22]. The ability of networks utilizing attention to learn arbitrary, non-local relationships has enabled a new state-of-the-art in NLP tasks, and attention-based architectures continue to be extended into areas historically dominated by convolutional networks such as image recognition [2, 3]. Furthermore, by applying attention mechanisms to transfer information from arbitrarily-shaped inputs into a fixed latent space, the application of attention-based 'Perceiver' models has demonstrated these architectures' potential as a 'universal' model which can answer complex queries on different tasks such as image classification, video compression, image flow, and more [9, 10].

Most attention-based architectures employ the popular 'query-key-value' (QKV) attention mechanism, which uses three inputs to produce a single output. Matrix multiplication is carried out between queries ($Q$) and keys ($K$). This matrix is then scaled by the dimensionality of the keys ($d_k$) for numerical stability and its softmax taken to calculate a set of scores, which represent the relevance between a given query and key. Matrix multiplication of the scores and values ($V$) then produces the output of the attention mechanism (Eqn. 5).

$$Attention(Q, K, V) = softmax\left(\frac{QK^T}{\sqrt{d_k}}\right)V$$
(Eqn. 5)

We adapt QKV attention to the FHRR domain by entirely replacing the scoring process with the VSA's similarity metric. That is, for each symbol in the set of queries and keys, the similarity between these symbols is calculated. This creates a score matrix on the domain [-1, 1]. The matrix multiplication of these scores and the values represented in the complex domain is used to produce the output of our VSA attention mechanism (Eqn. 6). This operation is carried out using only similarity and matrix multiplication, avoiding the need for arbitrary scaling and a softmax.

$$VSA\ Attention(Q, K, V) = sim.(Q, K) \cdot \exp(i\pi V)$$
(Eqn. 6)

## 2.3 Self-Attentional Module

By combining skip connections and attention mechanisms into a single module, we demonstrate a self-attention based architecture adapted for processing FHRR symbols. This was done with the goal of providing a powerful, trainable architecture to allow for the mapping of vector-symbols between domains.

A self-attentional architecture takes a set of inputs distributed over a space – positional, temporal, or otherwise – and uses attention to learn the relevance between information present in different inputs. For instance, an NLP attention model will 'attend to' the relevance between words at different positions in a sentence.

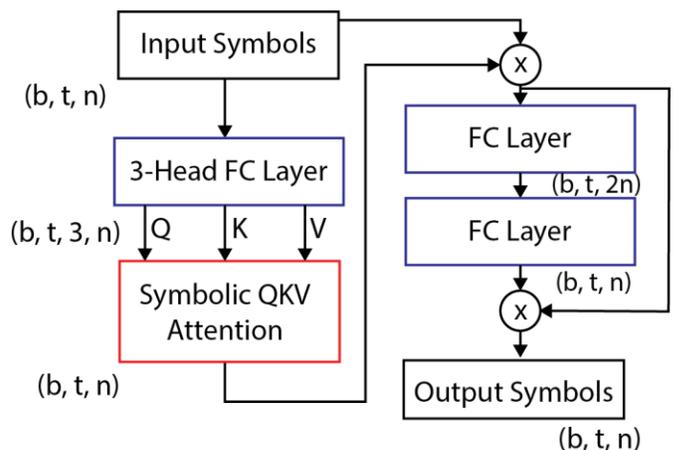

**Figure 4: Illustration of a self-attention module implemented for FHRR symbols. Neural layers are outlined in blue and the symbolic QKV attention is outlined in red. The VSA binding operation is indicated by circles around an 'x.' Fully-connected layers are labeled with 'FC' and use the generalized bundling function to compute outputs.**

To produce a VSA self-attention model, three fully-connected layers with an output size of $n$ (the dimensionality of the vector-symbols) are used to convert inputs to query, key, and value symbols. Symbolic QKV attention is then calculated and bound to the original inputs in a skip connection. This is followed by a residual block with two fully-connected layers (as used previously). The output of this block is bound with its inputs in a second skip connection. This produces the output of the symbolic self-attention block (Figure 4). All neural layers calculate outputs using the generalized bundling function, and losses through the block can be minimized via standard backpropagation.

## 2.4 Cross-Attentional Module

While self-attention blocks can theoretically be applied to any number of problems, the scaling of their computational footprint has previously prevented this in many cases. In a self-attention block, the score matrix scales with the number of input symbols squared. For large inputs such as those representing ImageNet images, this has prevented self-attention blocks from being directly applied to these problems [3, 9].

Cross-attention addresses this scaling issue by computing queries and keys/values from different sources. In a cross-attentional module, keys and values are produced directly from an input, but queries are instead produced from a fixed, trainable set which are known as 'inducing points' [14]. This rearrangement still allows for effective training of an attentional network while allowing the computation to scale linearly with the number of inputs.

By modifying the previous VSA self-attention module to produce keys/values from the inputs and queries from a trainable set, we construct a symbolic cross-attentional module (Figure 5).

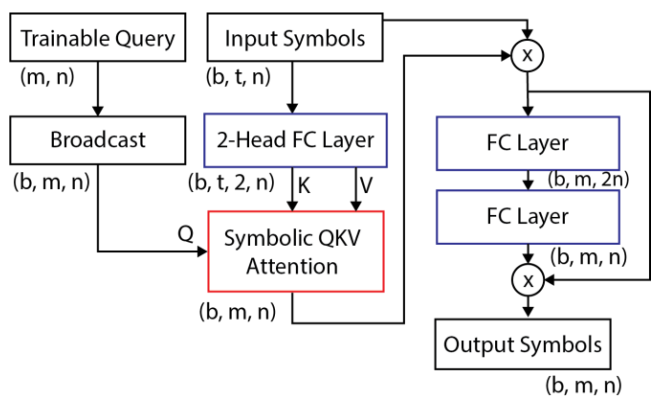

**Figure 5: Illustration of a cross-attention module implemented for FHRR symbols. This module is identical to the previous self-attention module with the exception of the query values, which are not produced from the inputs but are a set of trainable 'inducing points.'**

## 2.5 Image Classification

To test and validate our approach on a common task, we project individual rows of a FashionMNIST image into FHRR symbols, again using a Gaussian random projection and Layer Norm (Figure 6a). A self-attention block produces another set of FHRR symbols from these inputs, which are then reduced via a fully-connected layer into a single symbol. This symbol is then compared against a fixed 'codebook' which stores ten symbols, with each symbol representing one possible class. The class with the highest similarity to the output symbol is chosen as the network's class prediction (Figure 6c).

Alternately, a cross-attention block may replace the self-attention block. In this case, a set of trainable query values are used and images are only applied to produce the key/value inputs to the cross-attention module (Figure 6d). Otherwise, the architecture remains the same. For both architectures, training minimizes the distance between the model's output symbol and its matching class symbol (Figure 7).

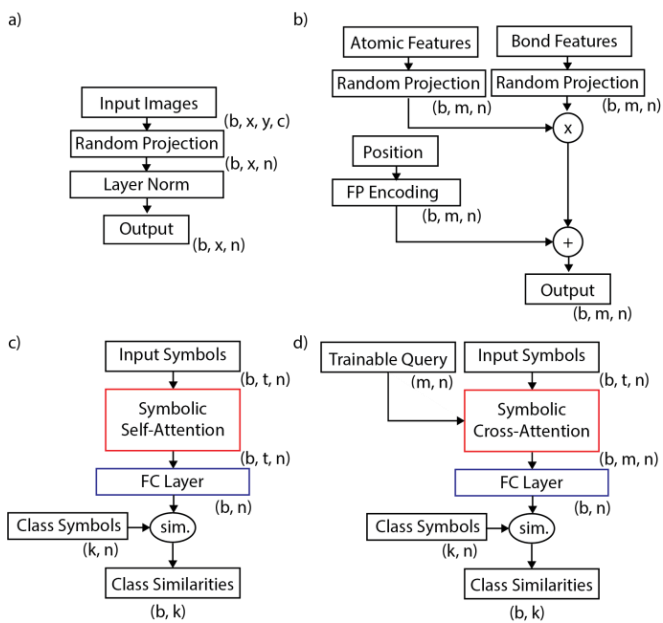

**Figure 6: Illustration of the architectures used for image classification and molecular toxicity prediction. In both problems, the input data – an image (a) or molecule (b) – is converted into a series of vector-symbols. These inputs can then be processed by domain-agnostic symbolic attention networks based on self-attention (c) or cross-attention (d).**

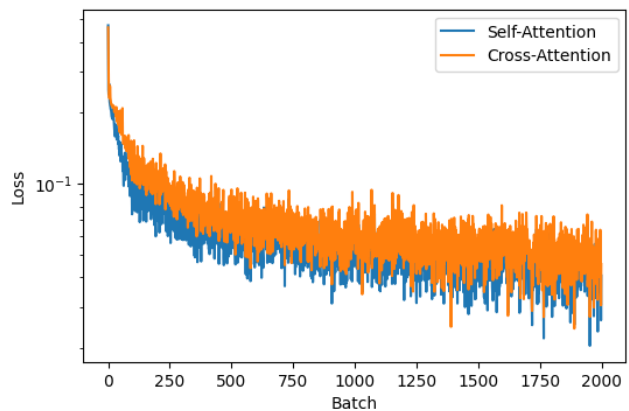

**Figure 7: Losses over training for symbolic self and cross-attention models classifying Fashion MNIST images. The cross-attentional model uses 32 trainable query symbols.**

To successfully predict a class, the attention module must learn to attend between a set of symbols, each of which represents a row from the original image. Both the self-attentional and cross-attentional architectures learn to do this effectively, reaching classification performance on the test set of 88.6% and 85.5%, respectively.

## 2.6 Drug Toxicity Prediction

Testing on FashionMNIST validates our approach on a simple task but does not leverage the ability of VSAs to compose and represent complex objects or demonstrate that symbolic attentional architectures can address problems in different domains. To address this, we apply the same attentional architectures used for FashionMNIST classification towards processing molecular structures provided by the CardioTox dataset [6]. This dataset includes characteristics of molecules, such as atoms and bonds, with the goal of using this information to predict whether the molecule can potentially bind with hERG, a protein involved in human heart activity. To do this, a prediction system must take a representation of the molecule as a graph and produce a prediction of toxicity with a confidence level.

In this dataset, each example consists of a molecule described by a graph, where nodes represent atoms and edges represent bonds. Both atoms and bonds contain a number of features. Each atom and bond's features are randomly projected into the FHRR domain to create symbols representing them. Symbols representing the two atoms involved in a bond and the bond's characteristics are bound to create a symbol representing each edge in the graph. These symbols are then bundled with a fractional power encoding representing position [11] to create a unique symbol for each edge in the graph (Figure 6b). This set of symbols is used to represent the molecule whose toxicity is being predicted. This set of symbolic inputs varies in length with the number of bonds in the input molecule. To enable batch-based processing, these inputs are padded with zeros to create a constant shape and applied as inputs into the same self and cross-attentional networks used for image classification (Figure 6c,d). In this case, the codebooks for these models only have two symbols – one represents 'toxic' and the other 'non-toxic.' Each network's final output symbol for each molecule is compared to this codebook, and the difference in absolute similarities between 'toxic' and 'non-toxic' is used as the model's confidence level for classification. Again, training minimizes the distance between the model's output symbol and the appropriate label symbol.

Performance on this test set is measured by area under receiving operator characteristic (AUROC) on three test sets. One test set is molecules which are similar to the training set (test-IID), and the others are molecules dissimilar to the training set (test-1, test-2). Results are summarized in Table 1. These models do not reach the level of state-of-the-art models [7], but do not require any domain-specific methods are initial results accomplished using a relatively simple graph encoding.

**Table 1: Results of attentional VSA model on molecular toxicity prediction vs. state-of-the-art model. All values represent area under the receiver operating characteristic (AUROC).**

| Network | Test-IID | Test-1 | Test-2 |
|---|---|---|---|
| Self-Attention, VSA | 0.86 | 0.71 | 0.59 |
| Cross-Attention, VSA | 0.84 | 0.71 | 0.67 |
| GNN-SNGP Ensemble [7] | 0.94 | 0.85 | 0.90 |

## 3 Discussion

In this work, we have focused on adapting techniques from deep learning to create network architectures suited for processing distributed, hyperdimensional symbols. We have aimed to do so using a limited set of operations which could conceivably be implemented via neuromorphic hardware: matrix multiplication in the complex domain and the FHRR operations of binding, bundling, and similarity. This novel approach can be extended in the future works by demonstrating that similarly to multi-layer perceptron models, spiking equivalents of these attentional architectures can be executed via the exchange of precisely-timed spikes [18].

In recent studies the Perceiver IO model [10] has demonstrated that combining self and cross-attentional modules can allow for designing advanced models for a variety of tasks. Key to the Perceiver IO model is the use of a specialized output query which can be constructed to specify tasks, such as optical flow at a given point in a video frame. Our proposed approach can potentially replicate the full Perceiver architecture using VSA attention modules and inputs/queries constructed via symbolic operations to firmly establish the parallels between these models and potentially enable their execution via neuromorphic hardware.

## 4 Conclusion

We have developed and demonstrated effective methods for implementing residual and attention based neural networks using only operations which are already required to compute with a specific Vector-Symbolic Architecture (VSA), the Fourier Holographic Reduced Representation (FHRR). This was done with the goal of providing novel and powerful methods for converting between different domains of symbolic or real information using operations compatible with hardware designed for VSA-based processing.

## 5 Methods

Data processing was carried out using Tensorflow 2.8.0 and all models were built using JAX 0.3.0 with Haiku 0.0.6. Code for models and experiments is publicly available at https://github.com/wilkieolin/FHRR_networks.

## ACKNOWLEDGMENTS

Acknowledgement. This work was supported by NIH T-32 Training Grant (5T32MH020002), the Lifelong Learning Machines program from DARPA/MTO (HR0011-18-2-0021), and ONR (MURI: N00014-16-1-2829).

# REFERENCES

[1] Ba, J.L. et al. 2016. Layer Normalization. (2016).

[2] Devlin, J. et al. 2019. BERT: Pre-training of deep bidirectional transformers for language understanding. *NAACL HLT 2019 - 2019 Conference of the North American Chapter of the Association for Computational Linguistics: Human Language Technologies - Proceedings of the Conference.* 1, Mlm (2019), 4171–4186.

[3] Dosovitskiy, A. et al. 2020. An Image is Worth 16x16 Words: Transformers for Image Recognition at Scale. (2020).

[4] Frady, E. et al. 2019. A framework for linking computations and rhythm-based timing patterns in neural firing, such as phase precession in hippocampal place cells. (2019). DOI:https://doi.org/10.32470/ccn.2018.1263-0.

[5] Goodfellow, I. et al. 2016. *Deep Learning*. MIT Press.

[6] Han, K. et al. 2021. Reliable Graph Neural Networks for Drug Discovery Under Distributional Shift. (2021).

[7] Han, K. et al. 2021. Reliable Graph Neural Networks for Drug Discovery Under Distributional Shift. (2021).

[8] He, K. et al. 2016. Deep residual learning for image recognition. *Proceedings of the IEEE Computer Society Conference on Computer Vision and Pattern Recognition.* 2016-Decem, (2016), 770–778. DOI:https://doi.org/10.1109/CVPR.2016.90.

[9] Jaegle, A. et al. 2021. Perceiver: General Perception with Iterative Attention. (2021).

[10] Jaegle, A. et al. 2021. Perceiver IO: A General Architecture for Structured Inputs & Outputs. (2021).

[11] Kleyko, D. et al. 2021. A Survey on Hyperdimensional Computing aka Vector Symbolic Architectures, Part I: Models and Data Transformations. (2021), 1–27.

[12] Kleyko, D. et al. 2021. A Survey on Hyperdimensional Computing aka Vector Symbolic Architectures, Part II: Applications, Cognitive Models, and Challenges. (2021), 1–36.

[13] Kleyko, D. et al. 2021. Vector Symbolic Architectures as a Computing Framework for Nanoscale Hardware. (2021), 1–28.

[14] Lee, J. et al. 2018. Set Transformer: A Framework for Attention-based Permutation-Invariant Neural Networks. (2018).

[15] Liu, W. et al. 2017. A survey of deep neural network architectures and their applications. *Neurocomputing.* 234, November 2016 (2017), 11–26. DOI:https://doi.org/10.1016/j.neucom.2016.12.038.

[16] Neubert, P. et al. 2019. An Introduction to Hyperdimensional Computing for Robotics. *KI - Künstliche Intelligenz.* 33, 4 (2019), 319–330. DOI:https://doi.org/10.1007/s13218-019-00623-z.

[17] Olin-Ammentorp, W. and Bazhenov, M. 2021. Bridge Networks: Relating Inputs through Vector-Symbolic Manipulations. *ICONS 2021* (2021).

[18] Olin-Ammentorp, W. and Bazhenov, M. 2021. Deep Phasor Networks: Connecting Conventional and Spiking Neural Networks. *arXiv.* 16 (2021).

[19] Plate, T.A. 2003. *Holographic Reduced Representation: Distributed Representation for Cognitive Structures.* Center for the Study of Language and Information.

[20] Plate, T.A. 1995. Holographic reduced representations. *IEEE Transactions on Neural networks.* 6, 3 (1995), 623–641.

[21] Schlegel, K. et al. 2020. A comparison of Vector Symbolic Architectures. (2020).

[22] Vaswani, A. et al. 2017. Attention is all you need. *Advances in Neural Information Processing Systems.* 2017-Decem, Nips (2017), 5999–6009.

[23] Xiao, H. et al. 2017. Fashion-MNIST: A novel image dataset for benchmarking machine learning algorithms. *arXiv.*